# CONVERGENCE OF CONTRASTIVE DIVERGENCE ALGORITHM IN EXPONENTIAL FAMILY

By Bai Jiang, Tung-Yu Wu, Yifan Jin and Wing H. Wong[*]

*Stanford University*

The Contrastive Divergence (CD) algorithm has achieved notable success in training energy-based models including Restricted Boltzmann Machines and played a key role in the emergence of deep learning. The idea of this algorithm is to approximate the intractable term in the exact gradient of the log-likelihood function by using short Markov chain Monte Carlo (MCMC) runs. The approximate gradient is computationally-cheap but biased. Whether and why the CD algorithm provides an asymptotically consistent estimate are still open questions. This paper studies the asymptotic properties of the CD algorithm in canonical exponential families, which are special cases of the energy-based model. Suppose the CD algorithm runs $m$ MCMC transition steps at each iteration $t$ and iteratively generates a sequence of parameter estimates $\{\theta_t\}_{t \geq 0}$ given an i.i.d. data sample $\{X_i\}_{i=1}^n \sim p_{\theta_\star}$. Under conditions which are commonly obeyed by the CD algorithm in practice, we prove the existence of some bounded $m$ such that any limit point of the time average $\sum_{s=0}^{t-1} \theta_s / t$ as $t \to \infty$ is a consistent estimate for the true parameter $\theta_\star$. Our proof is based on the fact that $\{\theta_t\}_{t \geq 0}$ is a homogenous Markov chain conditional on the data sample $\{X_i\}_{i=1}^n$. This chain meets the Foster-Lyapunov drift criterion and converges to a random walk around the Maximum Likelihood Estimate. The range of the random walk shrinks to zero at rate $\mathcal{O}(1/\sqrt[3]{n})$ as the sample size $n \to \infty$.

## 1. Introduction.

1.1. *Exponential Family and Maximum Likelihood Learning.* Consider a canonical exponential family over the sample space $\mathcal{X} \subseteq \mathbb{R}^p$ with the parameter $\theta \in \Theta \subseteq \mathbb{R}^d$

$$p_\theta(x) = c(x)e^{\theta^T \phi(x) - \Lambda(\theta)}, \tag{1.1}$$

where $c(x)$ is the carrier measure, $\phi : \mathcal{X} \to \mathbb{R}^d$ is the sufficient statistic and $\Lambda(\theta)$ is the cumulant generating function

$$\Lambda(\theta) := \log \int_\mathcal{X} c(x)e^{\theta^T \phi(x)} dx.$$

---

$\Lambda(\theta)$ is convex and differentiable at any interior point of the natural parameter domain. Denote by $\nabla\Lambda(\theta)$ and $\nabla^2\Lambda(\theta)$ the gradient vector and the Hessian matrix of $\Lambda(\theta)$, respectively. They are the expectation and the covariance of the sufficient statistic $\phi(X)$ under $p_\theta$. That is,

$$\text{(1.2)} \quad \nabla\Lambda(\theta) = \mathbb{E}_\theta \phi(X) = \int_\mathcal{X} \phi(x) p_\theta(x) dx,$$

$$\text{(1.3)} \quad \nabla^2\Lambda(\theta) = \text{Cov}_\theta \phi(X) = \int_\mathcal{X} [\phi(x) - \nabla\Lambda(\theta)][\phi(x) - \nabla\Lambda(\theta)]^T p_\theta(x) dx,$$

Given an i.i.d. sample $\mathbf{X} = \{X_i\}_{i=1}^n$ following a certain underlying distribution $p_{\theta_\star}$, the log-likelihood function is given by

$$l(\theta) := \frac{1}{n}\sum_{i=1}^n \log p_\theta(X_i) = \frac{1}{n}\sum_{i=1}^n \log c(X_i) + \theta^T \left[\frac{1}{n}\sum_{i=1}^n \phi(X_i)\right] - \Lambda(\theta).$$

Denote by $g(\theta)$ the gradient of $l(\theta)$.

$$g(\theta) := \nabla l(\theta) = \frac{1}{n}\sum_{i=1}^n \phi(X_i) - \nabla\Lambda(\theta).$$

The concavity of $l(\theta)$ follows from the convexity of $\Lambda(\theta)$. Successively iterating the update equation (1.4) of the gradient ascent algorithm will generate a sequence $\{\theta_t\}_{t\geq 0}$ indexed by the iteration number $t$. This sequence converges to the Maximum Likelihood Estimate (MLE) $\hat{\theta}_n := \text{argmax}_\theta\, l(\theta)$ if the learning rate $\eta$ is suitably chosen.

$$\text{(1.4)} \qquad \theta^+ = \theta + \eta g(\theta) = \theta + \eta \left[\frac{1}{n}\sum_{i=1}^n \phi(X_i) - \nabla\Lambda(\theta)\right].$$

In many cases, the term $\nabla\Lambda(\theta)$, essentially an integral as the expectation of the sufficient statistic in light of (1.2), is neither available in a simple closed form nor computationally tractable due to the complexity of the sample space $\mathcal{X}$ and/or the sufficient statistic $\phi$. An example is the fully-visible Boltzmann Machine (FVBM) model [1]. Its probability mass function is given by

$$\text{(1.5)} \qquad p_{W,b}(x) \propto \exp\left(\frac{1}{2}x^T W x + b^T x\right),$$

where $x$ is a $p$-dimensional vector of binary variables being either $-1$ or $+1$, $W$ is a $p \times p$ symmetric matrix with zero diagonal entries called "weight matrix" and $b$ is a $p$-dimensional vector called "bias vector". This model (1.5) is indeed an exponential family (1.1) with

$$\mathcal{X} = \{-1, +1\}^p,$$
$$\phi(x) = (x_i x_j, 1 \leq i < j \leq p; x_i, 1 \leq i \leq p) \in \mathbb{R}^{p(p+1)/2},$$
$$\theta = (W_{ij}, 1 \leq i < j \leq p; b_i, 1 \leq i \leq p) \in \mathbb{R}^{p(p+1)/2},$$
$$\Lambda(\theta) = \log \sum_{x \in \mathcal{X}} e^{\theta^T \phi(x)},$$

and $c(x)$ being the counting measure on the sample space $\mathcal{X}$. The gradient of the cumulant generating function

$$\nabla \Lambda(\theta) = \frac{\sum_{x \in \{-1,+1\}^p} \phi(x) e^{\theta^T \phi(x)}}{\sum_{x \in \{-1,+1\}^p} e^{\theta^T \phi(x)}} = \sum_{x \in \{-1,+1\}^p} \phi(x) p_\theta(x)$$

involves sums of exponentially many terms, which are computationally prohibitive even for a moderately high-dimensional $x$.

Markov Chain Monte Carlo (MCMC) is the standard approach to approximate $\nabla \Lambda(\theta)$. An MCMC run takes a large number of transition steps to reach the equilibrium, and gradient ascent algorithms iterate the update equation (1.4) hundreds or thousands of times. Thus it is computationally costly to implement a long MCMC run at each iteration of the gradient ascent algorithm (1.4).

1.2. *Contrastive Divergence.* In an influential paper [2], Hinton attempted to alleviate the long MCMC run time by first doing just a small number (say $m = 1, 2$ or 3) of transitions from the data sample $\{X_i\}_{i=1}^n$ as the initial values of the MCMC chains and then using the $m$-step MCMC sample $\{X_i^{(m)}\}_{i=1}^n$ to approximate $\nabla \Lambda(\theta)$. This is known as Hinton's Contrastive Divergence algorithm, hereafter abbreviated as CD, or CD-$m$ to also specify the fixed number $m$ of transitions. Formally, denote by $k_\theta(x, y)$ the MCMC transition kernel for the equilibrium distribution $p_\theta$. The CD-$m$ algorithm first runs $n$ Markov chains from $\{X_i\}_{i=1}^n$ independently for $m$ steps

$$X_1 \xrightarrow{k_\theta} X_1^{(1)} \xrightarrow{k_\theta} X_1^{(2)} \ldots \xrightarrow{k_\theta} X_1^{(m)}$$
$$X_2 \xrightarrow{k_\theta} X_2^{(1)} \xrightarrow{k_\theta} X_2^{(2)} \ldots \xrightarrow{k_\theta} X_2^{(m)}$$
$$\vdots$$
$$X_n \xrightarrow{k_\theta} X_n^{(1)} \xrightarrow{k_\theta} X_n^{(2)} \ldots \xrightarrow{k_\theta} X_n^{(m)}$$

and then uses $\{X_i^{(m)}\}_{i=1}^n$ to approximate $\nabla\Lambda(\theta)$ in the update equation (1.4) with

$$\nabla\Lambda(\theta) \approx \frac{1}{n}\sum_{i=1}^n \phi\left(X_i^{(m)}\right).$$

To sum up, the CD-$m$ algorithm replaces $g(\theta)$ in the update equation (1.4) of the gradient ascent algorithm with the CD gradient approximation

$$g_{\text{cd}}(\theta) \coloneqq \frac{1}{n}\sum_{i=1}^n \phi(X_i) - \frac{1}{n}\sum_{i=1}^n \phi\left(X_i^{(m)}\right)$$

and iterates the following update equation

$$(1.6) \qquad \theta^+ = \theta + \eta g_{\text{cd}}(\theta) = \theta + \eta\left[\frac{1}{n}\sum_{i=1}^n \phi(X_i) - \frac{1}{n}\sum_{i=1}^n \phi\left(X_i^{(m)}\right)\right].$$

The CD algorithm has been widely used by machine learners to train energy-based models of the form $p_\theta(x) = e^{-E(x,\theta)-\Lambda(\theta)}$, where $E(x,\theta)$ is called "energy function" and $\Lambda(\theta)$ is called "log-partition function". A notable example of these energy-based models is the Restricted Boltzmann Machines (RBM), which is the building block on each layer of deep belief network. The CD algorithm has performed well in layer-wise RBM pretraining stage of deep belief network and in this way played an key role in the emergence of deep learning [3, 4, 5, 6]. Applications of CD and RBM also include collaborative filtering [7], classification [8], topic modeling [9] and feature learning [10]. Apart from RBM, the CD algorithm has found practical applications in training energy-based models for acoustic modeling [11], image modeling [12, 13] and coarse-grained protein forcefield learning [14]. Exponential families (1.1) are special cases of energy-based models with $E(x,\theta) = -\theta^T\phi(x) - \log c(x)$. [15, 16, 17] use the CD algorithm to approximate MLE for exponential-family random graph models.

1.3. *Theoretical Studies on Contrastive Divergence.* There are a few open questions concerning the theoretical properties of the CD algorithm. First is whether or under what conditions the sequence $\{\theta_t\}_{t\geq 0}$ generated by the CD algorithm given a data sample $\mathbf{X} = \{X_i\}_{i=1}^n$ converges to some limit points as $t \to \infty$. If the answer is yes, we can regard these limit points as the CD estimates. Next questions of interest are how close these CD estimates are to the MLE, and whether they are asymptotically consistent for the true parameter $\theta_\star$ as $n \to \infty$.

Many eminent scholars in machine learning have attempted to answer these questions. MacKay [18] provided examples in which the CD-1 algorithm does not converge to the MLE as the iteration number $t \to \infty$. The reason is arguably the bias of the CD gradient approximation. The CD gradient has been proven biased in many models like Gaussian Boltzmann Machine and RBM, and the bias tends to decrease as $m$ increases in simulation studies [19, 20, 21, 22]. Yuille [23] stated formal conditions for the CD algorithm to converge to the true parameter (rather than the MLE) as $t \to \infty$. But he did not clearly distinguish the behavior of $\{\theta_t\}_{t \geq 0}$ in the limits of the iteration number $t \to \infty$ from that of the sample size $n \to \infty$, thus his conditions are not satisfied in even the simplest examples such as a bivariate Gaussian model (see more discussion in Section 5). For the FVBM model (1.5), Hyvärinen [24] shows that a specific CD-1 algorithm with the random-scan Gibbs sampler as the MCMC transition kernel is a stochastic version of an maximum pseudo-likelihood learning process, as the conditional expectation of the CD-1 gradient approximation given the data sample $\{X_i\}_{i=1}^n$ and the current parameter estimate $\theta$ is in the direction of the gradient of a certain pseudo-likelihood function. In this way, the author gave a heuristic argument for the consistency of the CD-1 algorithm in the specific setting. However, as the conditional expectation is not actually used in his CD-1 algorithm, this connection to the maximum pseudo-likelihood estimate cannot directly establish the consistency of the CD-1 algorithm.

This paper is devoted to answer the open question whether and why the CD algorithm with some bounded $m$ can yield an asymptotically consistent estimate. We restrict our focus to exponential families rather than general energy-based models for two reasons. First, the convexity of the cumulant generating function $\Lambda(\theta)$ in an exponential family guarantees the uniqueness of the MLE, enabling a transparent comparison of the CD algorithm to the maximum likelihood learning. The bias of the CD gradient approximation and the comparison of the CD estimates and MLE are of primary interest, because the idea of the CD algorithm is to replace the exact gradient with a computationally-cheap but biased CD gradient approximation when doing maximum (log-)likelihood learning. Second, the exponential family itself is a central statistical model. Yet except in the special case when the cumulant generating function is analytically tractable, there is no estimation method known to be asymptotically consistent and computationally efficient. As the CD algorithm appear to be a solution to these cases [15, 16, 17], its consistency for exponential families is of importance.

1.4. *Organization of Paper.* In practice, the CD algorithm iterates the update equation (1.6) many times ($t \to \infty$) to obtain an estimate given a particular data sample of size $n$. Thus, we first study the behavior of $\{\theta_t\}_{t \geq 0}$ in the limit of $t \to \infty$ given a data sample of fixed size $n$, and then let the sample size $n \to \infty$. The details of our approach can be stated as follows.

Conditional on a data sample of size $n$, the sequence $\{\theta_t\}_{t \geq 0}$ is a homogenous Markov chain. This chain has two phases: "quick move" and "random walk". When the chain moves from $\theta$ which is far away from the MLE $\hat{\theta}_n$, the exact gradient $g(\theta)$ is relatively large compared to the approximation error resulting from the $m$-step MCMC sampling. The update equation (1.6) keeps pushing $\theta$ to quickly move towards $\hat{\theta}_n$. When $\theta$ is so close to $\hat{\theta}_n$ that $g(\theta)$ fails to suppress the MCMC approximation error, the "quick move" phase ends and the chain starts a "random walk" in the neighborhood around $\hat{\theta}_n$. This intuition is mathematically formalized as a *Foster-Lyapunov drift criterion* with $V(\theta) = \|\theta - \hat{\theta}_n\|^2/2$ as a *Foster-Lyapunov function* (see Definition 3.2), where $\|z\|$ denotes the $l_2$-norm of vector $z$. This idea of a quick move to a random walk neighborhood and different types of Lyapunov drift conditions have been intensively explored in Markov chain theory [25, 26, 27]. Finally we show that the random walk neighborhood centering at $\hat{\theta}_n$ shrinks to the true parameter $\theta_\star$ as the sample size $n \to \infty$.

Section 2 states our main result: under six conditions (A1), (A2), (A3), (A4), (A5), and (A6), there exists a bounded $m$ for which the limiting time average $\frac{1}{t}\sum_{s=0}^{t-1}\theta_t$ converges to the true parameter $\theta_\star$ in probability

$$\limsup_{t \to \infty} \left\| \frac{1}{t}\sum_{s=0}^{t-1} \theta_s - \theta_\star \right\| \xrightarrow{p} 0$$

as the sample size $n \to \infty$ at a rate of $1/\sqrt[3]{n}$. Section 3 presents our proof in several stages. First, we show that $\{\theta_t\}_{t \geq 0}$ is a homogenous Markov chain under $\mathbb{P}^\mathbf{x}$, the conditional probability measure given any realization of the data sample $\mathbf{X} = \mathbf{x}$, and impose three constraints on $\mathbf{x}$ (and its sample size $n$). These constraints are proven to hold with probability approaching 1 as $n \to \infty$. Hereafter we study the chain $\{\theta_t\}_{t \geq 0}$ under $\mathbb{P}^\mathbf{x}$ in the framework of Markov chain and super-martingale theories and demonstrate that a neighborhood around the MLE $\hat{\theta}_n$ is positively recurrent. The key is to establish the Foster-Lyapunov drift criterion with $V(\theta) = \|\theta - \hat{\theta}_n\|^2/2$ as a Foster-Lyapunov function. From the Foster-Lyapunov drift criterion it follows that

$$\limsup_{t \to \infty} \left\| \frac{1}{t}\sum_{s=0}^{t-1} \theta_s - \hat{\theta}_n \right\| = \mathcal{O}(1/\sqrt[3]{n}) \quad \mathbb{P}^\mathbf{x}\text{-a.s.}.$$

Putting this $\mathbb{P}^{\mathbf{x}}$-a.s. convergence result and the fact that $\hat{\theta}_n$ is $\mathcal{O}_p(1/\sqrt{n})$-close to $\theta_\star$ together yields

$$\limsup_{t\to\infty} \left\| \frac{1}{t}\sum_{s=0}^{t-1} \theta_s - \theta_\star \right\| \leq \limsup_{t\to\infty} \left\| \frac{1}{t}\sum_{s=0}^{t-1} \theta_s - \hat{\theta}_n \right\| + \|\hat{\theta}_n - \theta_\star\| = \mathcal{O}_p(1/\sqrt[3]{n}).$$

Section 4 use a bivariate Gaussian model, an FVBM model (1.5) and a exponential-family random graph model (ERGM) as examples to illustrate the theories. Section 5 briefly discusses related works, novelties of our paper in theoretical aspects and guidance to practitioners of the CD algorithm.

**2. Main Result.** We base the asymptotic properties of the CD algorithm in canonical exponential families[1] on the assumptions (A1), (A2), (A3), (A4), (A5), and (A6). These assumptions can be directly verified in many applications of the CD algorithm in canonical exponential families. See three examples in Section 4.

(A1) The parameter space of interest $\Theta$ is a convex and compact subset of the natural parameter domain $\mathcal{D} = \{\theta \in \mathbb{R}^d : \Lambda(\theta) < \infty\}$, and the true parameter $\theta_\star$ is an interior point of $\Theta$.

The successive iterations of the update equation (1.6) may lead $\theta^+$ to leave compact $\Theta$. If it happens, we project $\theta^+$ onto $\Theta$. The remaining of this paper studies the modified update equation (2.1)

$$(2.1) \qquad \theta^+ = \Pi_\Theta \left( \theta + \eta \left[ \frac{1}{n}\sum_{i=1}^n \phi(X_i) - \frac{1}{n}\sum_{i=1}^n \phi\left(X_i^{(m)}\right) \right] \right),$$

where $\Pi_\Theta$ denotes the projection mapping onto $\Theta$ and is the proximal mapping associated to the convex function $h(\theta) = 0$ if $\theta \in \Theta$ or $\infty$ otherwise. This proximal mapping trick has been well studied by researchers focusing on the proximal gradient algorithm (see e.g. [28, 29]).

Let $\lambda_{\min}(\theta), \lambda_{\max}(\theta)$ be the smallest and largest eigenvalues of $\nabla^2 \Lambda(\theta)$, and let $\lambda_{\text{sum}}(\theta)$ be the trace (sum of eigenvalues) of $\nabla^2 \Lambda(\theta)$. The compactness of $\Theta$ in (A1) together with the positive definiteness and continuity of $\nabla^2 \Lambda(\theta)$ in a canonical exponential family imply the existence of the following

---

[1] An exponential family is canonical if the $d$-dimensional sufficient statistic $\phi(X)$ does not satisfy any linear constraint. If so, $\nabla^2 \Lambda(\theta) = \text{Cov}_\theta \phi(X)$ is positive definite.

constants

$$\lambda_{\min} := \inf_{\theta \in \Theta} \lambda_{\min}(\theta) \in (0, \infty), \tag{2.2}$$

$$\lambda_{\max} := \sup_{\theta \in \Theta} \lambda_{\max}(\theta) \in (0, \infty), \tag{2.3}$$

$$\lambda_{\text{sum}} := \sup_{\theta \in \Theta} \lambda_{\text{sum}}(\theta) \in (0, \infty) \tag{2.4}$$

We next explain how to quantify the difference of two distributions $p_\theta$ and $p_{\theta_\star}$ in an exponential family. We define $\chi^2$-contrast as follows. This $\chi^2$-contrast is commonly seen in the studies on the MCMC approximation error [30, 31].

DEFINITION 2.1 ($\chi^2$-contrast). *Let $\nu, \pi$ be two distributions on $\mathcal{X}$. If there exists a density of $\nu$ with respect to $\pi$ then denote it by $\frac{d\nu}{d\pi}(x)$. The $\chi^2$-contrast of $\nu$ and $\pi$ is given by*

$$\chi^2(\nu, \pi) = \int_{\mathcal{X}} \left[\frac{d\nu}{d\pi}(x) - 1\right]^2 \pi(x) dx = \int_{\mathcal{X}} \frac{[\nu(x) - \pi(x)]^2}{\pi(x)} dx.$$

*Let $\chi(\nu, \pi)$ be the square root of $\chi^2(\nu, \pi)$.*

(A2) There exists some positive constant $L$ such that

$$\chi(p_{\theta_\star}, p_\theta) \leq L\|\theta - \theta_\star\|, \ \forall \theta \in \Theta.$$

(A2) is not very restrictive. Indeed, the function $f : \theta \in \Theta \mapsto \chi(p_{\theta_\star}, p_\theta) = \sqrt{e^{-2\Lambda(\theta_\star) + \Lambda(\theta) + \Lambda(2\theta_\star - \theta)} - 1}$ is continuously differentiable and thus Lipschitz continuous in $\theta \in$ compact $\Theta$ as long as $\Lambda(2\theta_\star - \theta) < \infty$ for any $\theta \in \Theta$. Denote by $L$ the Lipschitz constant then

$$\chi(p_{\theta_\star}, p_\theta) = f(\theta) = f(\theta) - 0 = f(\theta) - f(\theta_\star) \leq L\|\theta - \theta_\star\|.$$

This condition holds in exponential families with $\Lambda(\theta) < \infty$ for any $\theta \in \mathbb{R}^d$. These exponential families include the FVBM model (1.5) and the Gaussian model with unknown mean $\theta$ and known covariance.

We also need a regularity condition on the MCMC transition kernel. Denote by $k_\theta(x, y)$ the MCMC transition kernel for the equilibrium distribution $p_\theta$, and by $K_\theta$ its associated Markov operator. That is, for any function $h : \mathcal{X} \to \mathbb{R}$,

$$K_\theta h(x) = \int_{\mathcal{X}} h(y) k_\theta(x, y) dy.$$

The Markov operator $K_\theta$ admits an $\mathcal{L}_2$-spectral gap, if

$$\alpha(\theta) := \sup_{h \neq 0} \left\{ \frac{\left[\int_\mathcal{X} |K_\theta h(x)|^2 p_\theta(x) dx\right]^{1/2}}{\left[\int_\mathcal{X} |h(x)|^2 p_\theta(x) dx\right]^{1/2}} : \int_\mathcal{X} h(x) p_\theta(x) dx = 0 \right\} < 1,$$

where the $\mathcal{L}_2$-spectral gap is given by $1 - \alpha(\theta)$. Larger $\mathcal{L}_2$-spectral gap indicates faster convergence rate of the MCMC chain [31]. (A3) requires that the Markov operators $\{K_\theta\}_{\theta \in \Theta}$ converge to their corresponding equilibrium $\{p_\theta\}_{\theta \in \Theta}$ uniformly fast.

(A3) The Markov operator $K_\theta$ admits an $\mathcal{L}_2$-spectral gap $1 - \alpha(\theta)$ and

$$\alpha := \sup_{\theta \in \Theta} \alpha(\theta) < 1.$$

Here we call $1 - \alpha$ the "uniform $\mathcal{L}_2$-spectral gap" of the Markov operators $\{K_\theta\}_{\theta \in \Theta}$. The spectrum theory for Markov operators has been elegantly established and studied [32, 33, 34]. This assumption is generally obeyed by popular MCMC transition kernels like Metropolis-Hastings algorithms and random-scan Gibbs samplers. These kernels usually generate reversible, $\varphi$-irreducible, and aperiodic chains in practice [35], and admit $\mathcal{L}_2$-spectral gaps if and only if they are geometrically ergodic (Proposition 1.2 in [36]). See [37, 38, 31] for more detailed discussions on $\mathcal{L}_2$-spectral gap and geometric ergodicity for MCMC algorithms. This "uniform $\mathcal{L}_2$-spectral gap" condition is equivalent to the "uniform geometric ergodicity" condition (H5) assumed in [39], which studies other MCMC-based estimation scheme.

Denote by $k_\theta^m(x, y)$ the $m$-step transition kernel

$$k_\theta^m(x, y) = \int_\mathcal{X} \cdots \int_\mathcal{X} k_\theta(x, x^{(1)}) k_\theta(x^{(1)}, x^{(2)}) \ldots k_\theta(x^{(m-1)}, y) dx^{(1)} \ldots dx^{(m-1)},$$

and by $k_\theta^m \nu$ the $m$-step transition of a (signed) measure $\nu$ on $\mathcal{X}$

$$k_\theta^m \nu(y) = \int_\mathcal{X} \nu(x) k_\theta^m(x, y) dx.$$

Then $k_\theta^m p_{\theta_\star}$ is the $m$-step transition of $p_{\theta_\star}$. In the CD algorithm, the $m$-step MCMC sample $\{X_i^{(m)}\}_{i=1}^n$ can in fact be regarded as i.i.d. draws from $k_\theta^m p_{\theta_\star}$. (A4) assumes that $\phi(X_i^{(m)})$ is sub-exponential.

DEFINITION 2.2 (sub-exponential random variable). *A $d$-dimensional random variable $Y$ is sub-exponential with parameters $(\sigma, \zeta)$ if*

$$\mathbb{E} e^{z^T(Y - \mathbb{E}Y)} \leq e^{\sigma^2 \|z\|^2 / 2}, \ \forall z \in \mathbb{R}^d \ s.t. \ \|z\| \leq \zeta.$$

When $d = 1$, the one-dimensional random variable $Y$ is said to be sub-exponential with parameters $(\sigma, \zeta)$ if

$$\mathbb{E}e^{z(Y-\mathbb{E}Y)} \leq e^{\sigma^2 z^2/2}, \ \forall z \in \mathbb{R} \ s.t. \ |z| \leq \zeta.$$

Apparently each component of a $d$-dimensional sub-exponential random variable is a one-dimensional sub-exponential random variable.

(A4) For any $\theta \in \Theta$, if $X^{(m)} \sim k_\theta^m p_{\theta_\star}$ then $\phi(X^{(m)})$ is sub-exponential with some constants $(\sigma_m, \zeta_m)$.

Sub-exponentiality is a commonly-seen condition in many statistics problems nowadays [40]. While many previous theoretical studies [23, 24] on the CD algorithm focused on cases with bounded $\phi(X^{(m)})$, this assumption covers the unbounded cases as long as the tail probability of $\phi(X^{(m)})$ is not heavy. Intuitively, (A4) is expected to holds as $\phi(X)$ is sub-exponential under both $p_\theta$ and $p_{\theta_\star}$ in regular exponential families (see Lemma 3.1 in the Appendix), and $k_\theta^m p_{\theta_\star}$ lies between the initial distribution $p_{\theta_\star}$ and the equilibrium distribution $p_\theta$. We also directly verify (A4) for the Gaussian, FVBM and ERGM examples in Section 4.

Let $k_\theta^m(x, \cdot)$ denote the $m$-step distribution of the chain starting from $x$. Then $X^{(m)}|x, \theta \sim k_\theta^m(x, \cdot)$. We assume that $\phi(X^{(m)})$ is conditionally square integrable given any $x \in \mathcal{X}$ and any $\theta \in \Theta$, and satisfies (A5) and (A6).

(A5) For any $\theta \in \Theta$, $f_\theta : x \mapsto \mathbb{E}\left[\phi(X^{(m)})|x;\theta\right] = \int_\mathcal{X} \phi(y) k_\theta^m(x,y) dy$ is a function of $x$. We assume that the mapping $\theta \mapsto f_\theta$ is Lipschitz continuous in the sense that there exists some positive constant $C_{1,m}$ (depending on $m$) such that

$$\sup_{x \in \mathcal{X}} \|f_{\theta_1}(x) - f_{\theta_2}(x)\| \leq C_{1,m}\|\theta_1 - \theta_2\|, \ \forall \theta_1, \theta_2 \in \Theta.$$

(A6) There exists some positive constant $C_{2,m}$ (depending on $m$) such that

$$\text{Cov}\left[\phi(X^{(m)})|x,\theta\right] \preceq C_{2,m} I_d, \ \forall \theta \in \Theta,$$

where $I_d$ denotes the $d \times d$ identity matrix, and $A \preceq B$ means $B - A$ is positive semi-definite for two symmetric matrices $A$ and $B$.

The intuition behind (A5) is that for two MCMC kernels using similar $\theta_1$ and $\theta_2$, the $m$-step transitions of the sufficient statistic $\phi(x)$ are similar. (A6) assumes the square integrability of $\phi(X^{(m)})$ under the $m$-step distribution $X^{(m)}|x;\theta \sim k_\theta^m(x,\cdot)$. They are commonly obeyed by the CD algorithm in practice. See examples in Section 4.

Now Theorem 2.1 states our main result.

THEOREM 2.1. *Assume (A1), (A2), (A3), (A4), (A5) and (A6). If the CD-m algorithm (2.1) generates a sequence $\{\theta_t\}_{t\geq 0}$ given an i.i.d. data sample $X_1, \ldots, X_n \sim p_{\theta_\star}$, then for any m and learning rate $\eta$ satisfying*

$$(2.5) \quad \lambda_{min} - \sqrt{\lambda_{sum}}L\alpha^m - \frac{\eta}{2}\left(\lambda_{max} + \sqrt{\lambda_{sum}}L\alpha^m\right)^2 > 0,$$

$$\lim_{n\to\infty} \mathbb{P}\left(\limsup_{t\to\infty} \left\|\frac{1}{t}\sum_{s=0}^{t-1}\theta_s - \theta_\star\right\| > A_m n^{-\gamma/3}\right) = 0$$

*for any $\gamma \in (0,1)$ and some constant $A_m$ depending on m. Here $\lambda_{min}, \lambda_{max}, \lambda_{sum}$ are defined by (2.2), (2.3) and (2.4), d is the dimension of $\theta$ and $\phi(x)$, L is the Lipschitz constant introduced by (A2), $1-\alpha$ is the uniform $\mathcal{L}_2$-spectral gap defined in (A3), and $\eta$ is the learning rate of the update equation (2.1).*

The left-hand side of (2.5) goes to $\lambda_{\min} > 0$ as $m \uparrow \infty$ and $\eta \downarrow 0$. There exist bounded m and $\eta$ to satisfy (2.5). For such m and $\eta$, the CD algorithm will give a consistent estimate for $\theta_\star$.

**3. Proof.** This section presents our proof in several stages. Subsection 3.1 shows that $\{\theta_t\}_{t\geq 0}$ is a homogenous Markov chain under $\mathbb{P}^{\mathbf{x}}$, the conditional probability measure given any realization of the data sample $\mathbf{X} = \mathbf{x}$, and imposes three constraints on $\mathbf{x}$ (and its sample size $n$). These constraints are proven to hold with probability approaching 1 as $n \to \infty$. The following subsections analyze the behaviors of the chain $\{\theta_t\}_{t\geq 0}$ under $\mathbb{P}^{\mathbf{x}}$ in the framework of Markov chain and super-martingale theories. Subsection 3.2 bounds the bias and the variance of the CD gradient approximation. With these bounds, we establish in Subsection 3.3 the Foster-Lyapunov drift criterion for the chain $\{\theta_t\}_{t\geq 0}$ with $V(\theta) = \|\theta - \hat{\theta}_n\|^2/2$ as a Foster-Lyapunov function. Subsection 3.4 follows to show that a neighborhood around the MLE $\hat{\theta}_n$ is positively recurrent, and further that

$$\limsup_{t\to\infty} \left\|\frac{1}{t}\sum_{s=0}^{t-1}\theta_s - \hat{\theta}_n\right\| = \mathcal{O}(1/\sqrt[3]{n}) \quad \mathbb{P}^{\mathbf{x}}\text{-a.s..}$$

Putting this $\mathbb{P}^{\mathbf{x}}$-a.s. convergence result and the fact that $\hat{\theta}_n$ is $\mathcal{O}_p(1/\sqrt{n})$-close to $\theta_\star$ together, Subsection 3.5 yields

$$\limsup_{t\to\infty}\left\|\frac{1}{t}\sum_{s=0}^{t-1}\theta_s - \theta_\star\right\| \leq \limsup_{t\to\infty}\left\|\frac{1}{t}\sum_{s=0}^{t-1}\theta_s - \hat{\theta}_n\right\| + \|\hat{\theta}_n - \theta_\star\| = \mathcal{O}_p(1/\sqrt[3]{n}).$$

3.1. *Conditioning on the Data Sample.* We claim that the CD algorithm (2.1) generates a homogenous Markov chain $\{\theta_t\}_{t\geq 0}$ in the state space $\Theta$ conditional on any realization of the data sample $\mathbf{X} = \mathbf{x}$.

Indeed, denote by $\mathbf{X}_t^{(m)} = \{X_{t,i}^{(m)}\}_{1\leq i\leq n}$ the $m$-step MCMC sample generated at iteration $t$ of the CD algorithm. The filtration

$$\mathcal{F}_t \coloneqq \sigma\text{-algebra}\left(\mathbf{X}, \theta_0, \mathbf{X}_1^{(m)}, \theta_1, \mathbf{X}_2^{(m)}, ..., \theta_{t-1}, \mathbf{X}_t^{(m)}, \theta_t\right)$$

contains all historical information until iteration $t$. At each iteration $t$, the CD update

$$\theta_t = \Pi_\Theta\left(\theta_{t-1} + \eta\left[\frac{1}{n}\sum_{i=1}^n \phi(x_i) - \frac{1}{n}\sum_{i=1}^n \phi\left(X_{t,i}^{(m)}\right)\right]\right)$$

is merely a function of the data sample $\mathbf{X} = \mathbf{x}$, the current parameter estimate $\theta_{t-1}$ and the $m$-step MCMC sample $\mathbf{X}_t^{(m)}$. Given the data sample $\mathbf{X} = \mathbf{x}$ and the current parameter estimate $\theta_{t-1}$, $\mathbf{X}_t^{(m)}$ is conditionally independent to the past history of CD updates. Thus $\{\theta_t\}_{t\geq 0}$ is a homogeneous $\mathcal{F}_t$-adapted Markov chain under $\mathbb{P}^\mathbf{x}$, the conditional probability measure given $\mathbf{X} = \mathbf{x}$.

Next, we impose three constraints (3.1), (3.2) and (3.3) on the data sample $\mathbf{X} = \mathbf{x}$. Lemma 3.1 proves that they hold with probability approaching 1 as $n \to \infty$. In the following sections, we study the chain $\{\theta_t\}_{t\geq 0}$ under $\mathbb{P}^\mathbf{x}$ with $\mathbf{x}$ satisfying these constraints.

LEMMA 3.1. *Assume (A1), (A4), (A5) and $X_1, ..., X_n \overset{i.i.d.}{\sim} p_{\theta_\star}$. Denote by $\partial\Theta$ the boundary of compact $\Theta$, by $\hat\theta_n$ the MLE, and by $f_\theta : x \mapsto \mathbb{E}\left[\phi(X^{(m)})|x, \theta\right]$ the function defined in (A5). For any $\gamma \in (0,1)$,*

$$\inf_{\theta \in \partial\Theta} \|\theta - \theta_\star\| > n^{-\gamma/2} \tag{3.1}$$

$$\|\hat\theta_n - \theta_\star\| < n^{-\gamma/2} \tag{3.2}$$

$$\sup_{\theta \in \Theta} \left\|\frac{1}{n}\sum_{i=1}^n f_\theta(X_i) - \mathbb{E}f_\theta(X_1)\right\| < n^{-\gamma/2} \tag{3.3}$$

*hold with probability approaching 1 as $n \to \infty$.*

PROOF. (3.1) holds for sufficiently large $n$ since the true parameter $\theta_\star$ is an interior point of compact $\Theta$ as assumed in (A1). The standard theorem for the MLE [41] asserts that (3.2) holds with probability approaching 1 as $n \to \infty$. Only left is to show (3.3) holds with probability approaching 1 as $n \to \infty$.

To this end, consider $N = \mathcal{O}(\epsilon^{-d})$ $\epsilon$-balls to cover $\Theta$, which center at $\{\theta_l\}_{1 \leq l \leq N}$. Any $\theta \in \Theta$ is $\epsilon$-close to at least one $\theta_l$. (A5) implies

$$\sup_{x \in \mathcal{X}} \|f_\theta(x) - f_{\theta_l}(x)\| \leq C_{1,m}\epsilon$$

and further

$$\left\| \frac{1}{n} \sum_{i=1}^n f_\theta(X_i) - \mathbb{E} f_\theta(X_1) \right\| \leq \left\| \frac{1}{n} \sum_{i=1}^n f_\theta(X_i) - \frac{1}{n} \sum_{i=1}^n f_{\theta_l}(X_i) \right\|$$
$$+ \left\| \frac{1}{n} \sum_{i=1}^n f_{\theta_l}(X_i) - \mathbb{E} f_{\theta_l}(X_1) \right\|$$
$$+ \|\mathbb{E} f_{\theta_l}(X_1) - \mathbb{E} f_\theta(X_1)\|$$
$$\leq \left\| \frac{1}{n} \sum_{i=1}^n f_{\theta_l}(X_i) - \mathbb{E} f_{\theta_l}(X_1) \right\| + 2C_{1,m}\epsilon.$$

It follows that

$$\mathbb{P}\left( \sup_{\theta \in \Theta} \left\| \frac{1}{n} \sum_{i=1}^n f_\theta(X_i) - \mathbb{E} f_\theta(X_1) \right\| \geq 3C_{1,m}\epsilon \right)$$
$$\leq \mathbb{P}\left( \max_{l=1}^N \left\| \frac{1}{n} \sum_{i=1}^n f_{\theta_l}(X_i) - \mathbb{E} f_{\theta_l}(X_1) \right\| \geq C_{1,m}\epsilon \right)$$
$$\leq \sum_{l=1}^N \mathbb{P}\left( \left\| \frac{1}{n} \sum_{i=1}^n f_{\theta_l}(X_i) - \mathbb{E} f_{\theta_l}(X_1) \right\| \geq C_{1,m}\epsilon \right)$$
$$\leq \sum_{l=1}^N \sum_{j=1}^d \mathbb{P}\left( \left| \frac{1}{n} \sum_{i=1}^n f_{\theta_l,j}(X_i) - \mathbb{E} f_{\theta_l,j}(X_1) \right| \geq \frac{C_{1,m}\epsilon}{\sqrt{d}} \right)$$

$\phi(X_i^{(m)})$ is sub-exponential with $(\sigma_m, \zeta_m)$ as assumed in (A4), so is its conditional expectation $f_\theta(X_i) = \mathbb{E}\left[\phi(X_i^{(m)})|X_i, \theta\right]$ by Lemma 3.2 in the Appendix. Let $f_{\theta,j}$ be the $j$-th component of $f_\theta$ then one-dimensional random variables $\{f_{\theta,j}(X_i)\}_{i=1}^n$ are i.i.d. and sub-exponential with $(\sigma_m, \zeta_m)$. By Lemma 3.3 in the Appendix,

$$\mathbb{P}\left( \left| \frac{1}{n} \sum_{i=1}^n f_{\theta_l,j}(X_i) - \mathbb{E} f_{\theta_l,j}(X_1) \right| \geq \frac{C_{1,m}\epsilon}{\sqrt{d}} \right) \leq 2 \exp\left( -\frac{nC_{1,m}^2 \epsilon^2 / d}{2\sigma_m^2} \right)$$

if $C_{1,m}\epsilon/\sqrt{d} < \sigma_m^2 \zeta_m$. Putting together with the fact that $N = \mathcal{O}(\epsilon^{-d})$ yields

$$\mathbb{P}\left(\sup_{\theta \in \Theta} \left\| \frac{1}{n} \sum_{i=1}^n f_\theta(X_i) - \mathbb{E}f_\theta(X_1) \right\| \geq 3C_{1,m}\epsilon \right) \leq 2Nd \exp\left(-\frac{nC_{1,m}^2 \epsilon^2/d}{2\sigma_m^2}\right)$$
$$= \mathcal{O}(\epsilon^{-d}) \exp\left(-\frac{nC_{1,m}^2 \epsilon^2/d}{2\sigma_m^2}\right)$$

if $C_{1,m}\epsilon/\sqrt{d} < \sigma_m^2 \zeta_m$. Let $\epsilon = n^{-\gamma/2}/3C_{1,m}$ then $C_{1,m}\epsilon/\sqrt{d} < \sigma_m^2 \zeta_m$ if $n$ is sufficiently large. Thus

$$\mathbb{P}\left(\sup_{\theta \in \Theta} \left\| \frac{1}{n} \sum_{i=1}^n f_\theta(X_i) - \mathbb{E}f_\theta(X_1) \right\| \geq n^{-\gamma/2} \right) \leq \mathcal{O}(n^{\gamma d/2}) \exp\left(-\frac{n^{1-\gamma}}{18 d \sigma_m^2}\right)$$
$$\to 0$$

as $n \to \infty$, completing the proof. $\square$

3.2. *Gradient Approximation Error.* Denote by $\mathbb{P}_\theta^{\mathbf{x}}$ the conditional probability measure given the data sample $\mathbf{x}$ and the current state $\theta_t = \theta$ of the chain. And $\mathbb{E}_\theta^{\mathbf{x}}$ and $\mathbb{C}\text{ov}_\theta^{\mathbf{x}}$ denote the expectation and covariance under $\mathbb{P}_\theta^{\mathbf{x}}$. Lemma 3.2 bounds the approximation error of the CD gradient $g_{\text{cd}}(\theta)$ under $\mathbb{P}_\theta^{\mathbf{x}}$. Specifically, the bias of $g_{\text{cd}}(\theta)$ is $\mathcal{O}(n^{-\gamma/2}) + \mathcal{O}(\alpha^m \|\theta - \hat{\theta}_n\|)$, which depends on the uniform $\mathcal{L}_2$-spectral gap $1 - \alpha$ of the MCMC kernels, the number $m$ of transition steps in MCMC, the sample size $n$ and the distance between $\theta$ and the MLE $\hat{\theta}_n$. This result agrees with Bengio and Delalleau [21]'s finding that the CD gradient approximation error decreases at a rate depending on the mixing rate of the MCMC kernels.

LEMMA 3.2. *Assume (A1), (A2), (A3), (A6), and that the data sample $\mathbf{x} = \{x_i\}_{i=1}^n$ satisfies (3.2) and (3.3). Let $\Delta g = g_{cd}(\theta) - g(\theta)$ be the gradient approximation error. Then*

$$\|\mathbb{E}_\theta^{\mathbf{x}} \Delta g\| \leq \left(1 + \sqrt{\lambda_{sum}} L \alpha^m\right) n^{-\gamma/2} + \sqrt{\lambda_{sum}} L \alpha^m \|\theta - \hat{\theta}_n\|$$

*where $\lambda_{sum}$ is defined in (2.4), $L$ denotes the Lipchitz constant introduced by (A2), $1 - \alpha$ is the uniform $\mathcal{L}_2$-spectral gap defined in (A3), and $\gamma \in (0, 1)$ is introduced by constraints (3.2) and (3.3). Also,*

$$\mathbb{C}\text{ov}_\theta^{\mathbf{x}} \Delta g \preceq \frac{C_{2,m}}{n} I_d,$$

*where $C_{2,m}$ is defined in (A6).*

PROOF. From the fact that

$$\Delta g = g_{\text{cd}}(\theta) - g(\theta) = \nabla \Lambda(\theta) - \frac{1}{n}\sum_{i=1}^n \phi(X_i^{(m)}),$$

it follows that

$$-\mathbb{E}^{\mathbf{x}}\Delta g = \mathbb{E}_\theta^{\mathbf{x}}\left[\frac{1}{n}\sum_{i=1}^n \phi(X_i^{(m)})\right] - \nabla\Lambda(\theta)$$

$[X_i^{(m)}|x_i, \theta \sim k_\theta^m(x_i, \cdot)]$
$$= \frac{1}{n}\sum_{i=1}^n \int_{\mathcal{X}} \phi(y)k_\theta^m(x_i, y)dy - \nabla\Lambda(\theta)$$

[Definition of $f_\theta$ in (A5)]
$$= \frac{1}{n}\sum_{i=1}^n f_\theta(x_i) - \nabla\Lambda(\theta)$$

$$= \left[\frac{1}{n}\sum_{i=1}^n f_\theta(x_i) - \mathbb{E}f_\theta(X_1)\right]$$
$$+ [\mathbb{E}f_\theta(X_1) - \nabla\Lambda(\theta)]$$

Constraint (3.3) bounds the length of the first term (vector) by $n^{-\gamma/2}$. Proceed to consider the second term $\mathbb{E}f_\theta(X_1) - \nabla\Lambda(\theta)$. Write

$[X_1 \sim p_{\theta_\star}]$
$$\mathbb{E}f_\theta(X_1) = \int_{\mathcal{X}} f_\theta(x)p_{\theta_\star}(x)dx$$

[Definition of $f_\theta$ in (A5)]
$$= \int_{\mathcal{X}}\left(\int_{\mathcal{X}} \phi(y)k_\theta^m(x,y)dy\right)p_{\theta_\star}(x)dx$$

[Fubini's theorem]
$$= \int_{\mathcal{X}} \phi(y)\left(\int_{\mathcal{X}} k_\theta^m(x,y)p_{\theta_\star}(x)dx\right)dy$$

[Definition of $k_\theta^m p_{\theta_\star}$]
$$= \int_{\mathcal{X}} \phi(y)k_\theta^m p_{\theta_\star}(y)dy.$$

The fact that

$$\nabla\Lambda(\theta) = \int_{\mathcal{X}} \nabla\Lambda(\theta)k_\theta^m p_{\theta_\star}(x)dx = \int_{\mathcal{X}} \nabla\Lambda(\theta)p_\theta(x)dx$$

and (1.2) imply

$$\mathbb{E}f_\theta(X_1) - \nabla\Lambda(\theta) = \int_{\mathcal{X}} \phi(x)k_\theta^m p_{\theta_\star}(x)dx - \int_{\mathcal{X}} \phi(x)p_\theta(x)dx$$
$$- \int_{\mathcal{X}} \nabla\Lambda(\theta)k_\theta^m p_{\theta_\star}(x)dx + \int_{\mathcal{X}} \nabla\Lambda(\theta)p_\theta(x)dx$$
$$= \int_{\mathcal{X}} [\phi(x) - \nabla\Lambda(\theta)][k_\theta^m p_{\theta_\star}(x) - p_\theta(x)]dx$$

For each $j = 1, ..., d$, let $f_{\theta,j}(x), \phi_j(x)$ and $\nabla_j \Lambda(\theta) = \partial \Lambda(\theta)/\partial \theta_j$ be the $j$-th component of $f_\theta(x), \phi(x)$ and $\nabla \Lambda(\theta)$, respectively. Let $\nabla^2_{jj}\Lambda(\theta) = \partial^2 \Lambda(\theta)/\partial \theta_j^2$ be the $j$-th diagonal entry of $\nabla^2 \Lambda(\theta)$.

$$\begin{aligned}
|\mathbb{E} f_{\theta,j}(X_1) - \nabla_j \Lambda(\theta)| &= \left| \int_{\mathcal{X}} [\phi_j(x) - \nabla_j \Lambda(\theta)][k_\theta^m p_{\theta_\star}(x) - p_\theta(x)]dx \right| \\
&= \left| \int_{\mathcal{X}} [\phi_j(x) - \nabla_j \Lambda(\theta)] \left[ \frac{k_\theta^m p_{\theta_\star}(x)}{p_\theta(x)} - 1 \right] p_\theta(x)dx \right| \\
[\text{Cauchy-Schwartz}] &\leq \sqrt{\int_{\mathcal{X}} [\phi_j(x) - \nabla_j \Lambda(\theta)]^2 p_\theta(x)dx} \\
&\quad \times \sqrt{\int_{\mathcal{X}} \left[ \frac{k_\theta^m p_{\theta_\star}(x)}{p_\theta(x)} - 1 \right]^2 p_\theta(x)dx} \\
[(1.3), \text{Definition } 2.1] &= \sqrt{\nabla^2_{jj}\Lambda(\theta)} \times \chi(k_\theta^m p_{\theta_\star}, p_\theta)
\end{aligned}$$

Noting that $\lambda_{\text{sum}}(\theta) = \text{trace}[\nabla^2 \Lambda(\theta)]$ and that $\chi(k_\theta^m p_{\theta_\star}, p_\theta) \leq \alpha(\theta)^m \chi(p_{\theta_\star}, p_\theta)$ due to Lemma 3.4 in the Appendix (also part of Theorem 2.1 in [30] and Proposition 3.12 in [31]), we further have

(3.4) $$\|\mathbb{E} f_\theta(X_1) - \nabla \Lambda(\theta)\| \leq \sqrt{\lambda_{\text{sum}}(\theta)} \times \alpha(\theta)^m \chi(p_{\theta_\star}, p_\theta)$$

Hence

$$\begin{aligned}
\|\mathbb{E}_\theta^{\mathbf{x}} \Delta g\| &\leq \left\| \frac{1}{n} \sum_{i=1}^n f_\theta(x_i) - \mathbb{E} f_\theta(X_1) \right\| + \|\mathbb{E} f_\theta(X_1) - \nabla \Lambda(\theta)\| \\
[(3.3), (3.4)] &\leq n^{-\gamma/2} + \sqrt{\lambda_{\text{sum}}(\theta)} \times \alpha(\theta)^m \chi(p_{\theta_\star}, p_\theta) \\
[(2.4), (A2), (A3)] & \\
&\leq n^{-\gamma/2} + \sqrt{\lambda_{\text{sum}}} L\alpha^m \|\theta_\star - \theta\| \\
&\leq n^{-\gamma/2} + \sqrt{\lambda_{\text{sum}}} L\alpha^m \|\hat{\theta}_n - \theta_\star\| + \sqrt{\lambda_{\text{sum}}} L\alpha^m \|\theta - \hat{\theta}_n\| \\
[(3.2)] &\leq \left(1 + \sqrt{\lambda_{\text{sum}}} L\alpha^m\right) n^{-\gamma/2} + \sqrt{\lambda_{\text{sum}}} L\alpha^m \|\theta - \hat{\theta}_n\|.
\end{aligned}$$

$X_i^{(m)}|\mathbf{x}, \theta \sim k_\theta^m(x_i, \cdot)$ are conditionally independent (but not identically distributed) since $n$ chains independently starts from different $x_i$. By (A6),

$$\mathbb{C}\text{ov}_\theta^{\mathbf{x}} \Delta g = \frac{1}{n^2} \sum_{i=1}^n \mathbb{C}\text{ov}\left[\phi(X_i^{(m)})|x_i, \theta\right] \preceq \frac{C_{2,m}}{n} I_d.$$

□

3.3. *Foster-Lyapunov drift criterion.* This subsection uses the error bounds of CD gradient approximation in Lemma 3.2 to establish the *Foster-Lyapunov drift criterion* with

$$V(\theta) := \|\theta - \hat{\theta}_n\|^2/2$$

as a *Foster-Lyapunov function*. See Definitions 3.1 and 3.2 for drift, Foster-Lyapunov drift criterion and Foster-Lyapunov function.

DEFINITION 3.1 (drift). *Let $V : \Theta \to \mathbb{R}_+$ be some non-negative function on the state space of a Markov chain $\{\theta_t\}_{t \geq 0}$. The one-step drift of $V$ is defined as $\mathbb{E}_\theta V(\theta^+) - V(\theta)$, which is the expected value change of $V$ when the chain moves from $\theta$ to $\theta^+$.*

DEFINITION 3.2 (Foster-Lyapunov drift criterion). *A Markov chain $\{\theta_t\}_{t \leq 0}$ satisfies the Foster-Lyapunov drift criterion if*

$$\mathbb{E}_\theta V(\theta^+) - V(\theta) \leq -\delta_1 \mathbb{I}(\theta \notin B) + \delta_2 \mathbb{I}(\theta \in B)$$

*with some $\delta_1, \delta_2 > 0$ and some subset $B$ of the state space $\Theta$. $V$ is called a Foster-Lyapunov function.*

Lemma 3.3 shows that the chain $\{\theta_t\}_{t \geq 0}$ satisfies the *Foster-Lyapunov drift criterion*.

LEMMA 3.3. *Assume (A1), (A2), (A3), (A6), and that the data sample $\mathbf{x}$ satisfies (3.1), (3.2) and (3.3) for some $\gamma \in (0, 1)$. Then for any $m$ and learning rate $\eta$ satisfying*

$$a := \lambda_{min} - \sqrt{\lambda_{sum}} L\alpha^m - \frac{\eta}{2}\left(\lambda_{max} + \sqrt{\lambda_{sum}} L\alpha^m\right)^2 > 0,$$

*the chain $\{\theta_t\}_{t \geq 0}$ satisfies Foster-Lyapunov drift criterion*

$$\mathbb{E}_\theta^{\mathbf{x}} V(\theta^+) - V(\theta) \leq -\delta_1 \mathbb{I}(\theta \notin B) + \delta_2 \mathbb{I}(\theta \in B),$$

*with the Foster-Lyapunov function*

$$V(\theta) = \|\theta - \hat{\theta}_n\|^2/2$$

*and*

$$B = \{\theta \in \Theta : \|\theta - \hat{\theta}_n\| \leq \beta r_n\}, \quad \delta_1 = \eta(\beta^2 - 1)c_n, \quad \delta_2 = \eta(c_n + b_n^2/4a).$$

Here $\beta > 1$ is arbitrary and $b_n, c_n, r_n$ are defined in the following way

$$b_n := \left(1 + \sqrt{\lambda_{sum}}L\alpha^m\right)\left(1 + \eta\lambda_{max} + \eta\sqrt{\lambda_{sum}}L\alpha^m\right)n^{-\gamma/2}$$

$$c_n := \frac{\eta}{2}\left[dC_{2,m}n^{-1+\gamma} + \left(1 + \sqrt{\lambda_{sum}}L\alpha^m\right)^2\right]n^{-\gamma}$$

$$r_n := \frac{b_n + \sqrt{b_n^2 + 4ac_n}}{2a} \asymp n^{-\gamma/2}$$

PROOF. Let $\Pi_\Theta$ denote the projection mapping onto $\Theta$. From (3.1) and (3.2) it follows that $\hat{\theta}_n \in \Theta$, further implying that

$$\hat{\theta}_n = \Pi_\Theta(\hat{\theta}_n).$$

Let $\Delta g = g_{\mathrm{cd}}(\theta) - g(\theta)$ be the approximation error of the CD gradient. We analyze the one-step drift of the update equation (2.1). Write

$$\begin{aligned}
V(\theta^+) &= \frac{1}{2}\|\theta^+ - \hat{\theta}_n\|^2 = \frac{1}{2}\|\Pi_\Theta(\theta + \eta g_{\mathrm{cd}}(\theta)) - \Pi_\Theta(\hat{\theta}_n)\|^2 \\
&\leq \frac{1}{2}\|\theta + \eta g_{\mathrm{cd}}(\theta) - \hat{\theta}_n\|^2 \\
&= V(\theta) + \eta(\theta - \hat{\theta}_n)^T g_{\mathrm{cd}}(\theta) + \frac{\eta^2}{2}\|g_{\mathrm{cd}}(\theta)\|^2 \\
&= V(\theta) + \eta(\theta - \hat{\theta}_n)^T g(\theta) + \eta(\theta - \hat{\theta}_n)^T\Delta g + \frac{\eta^2}{2}\|g_{\mathrm{cd}}(\theta)\|^2,
\end{aligned}$$

implying the one-step drift

$$\begin{aligned}
\mathbb{E}_\theta^{\mathbf{x}} V(\theta^+) - V(\theta) &\leq \eta(\theta - \hat{\theta}_n)^T g(\theta) + \eta(\theta - \hat{\theta}_n)^T\mathbb{E}_\theta^{\mathbf{x}}\Delta g + \frac{\eta^2}{2}\mathbb{E}_\theta^{\mathbf{x}}\left[\|g_{\mathrm{cd}}(\theta)\|^2\right] \\
&\leq \eta(\theta - \hat{\theta}_n)^T g(\theta) + \eta(\theta - \hat{\theta}_n)^T\mathbb{E}_\theta^{\mathbf{x}}\Delta g \\
&\quad + \frac{\eta^2}{2}\|\mathbb{E}_\theta^{\mathbf{x}} g_{\mathrm{cd}}(\theta)\|^2 + \frac{\eta^2}{2}\mathrm{trace}\left[\mathrm{Cov}_\theta^{\mathbf{x}}\Delta g\right] \\
&\leq \eta(\theta - \hat{\theta}_n)^T g(\theta) + \eta\|\theta - \hat{\theta}_n\|\|\mathbb{E}_\theta^{\mathbf{x}}\Delta g\| \\
&\quad + \frac{\eta^2}{2}\left(\|g(\theta)\| + \|\mathbb{E}_\theta^{\mathbf{x}}\Delta g\|\right)^2 + \frac{\eta^2}{2}\mathrm{trace}\left[\mathrm{Cov}_\theta^{\mathbf{x}}\Delta g\right].
\end{aligned}$$
(3.5)

From the facts that $g(\hat{\theta}_n) = 0$ and that $\nabla g(\theta) = -\nabla^2\Lambda(\theta)$, it follows that

$$g(\theta) = g(\theta) - g(\hat{\theta}_n) = -\nabla^2\Lambda(\theta')(\theta - \hat{\theta}_n)$$

for some $\theta'$ between $\theta$ and $\hat{\theta}_n$. In the first term of the right-hand side of (3.5),

$$\begin{aligned}
(\theta - \hat{\theta}_n)^T g(\theta) &= -(\theta - \hat{\theta}_n)^T\nabla^2\Lambda(\theta')(\theta - \hat{\theta}_n) \\
&\leq -\lambda_{\min}\|\theta - \hat{\theta}_n\|^2
\end{aligned}$$

In the third term of the right-hand side of (3.5),

$$\begin{aligned}
\|g(\theta)\| &= \|\nabla^2\Lambda(\theta')(\theta - \hat\theta_n)\| \\
&= \sqrt{(\theta - \hat\theta_n)^T [\nabla^2\Lambda(\theta')]^2 (\theta - \hat\theta_n)} \\
&\leq \sqrt{\lambda_{\max}^2 \|\theta - \hat\theta_n\|^2} \\
&= \lambda_{\max} \|\theta - \hat\theta_n\|
\end{aligned}$$

Plugging them and results in Lemma 3.2 together into (3.5) yields

$$\begin{aligned}
&\mathbb{E}_\theta^{\mathbf{x}} V(\theta^+) - V(\theta) \\
&\leq -\eta\lambda_{\min}\|\theta - \hat\theta_n\|^2 \\
&\quad + \eta\left[\left(1 + \sqrt{\lambda_{\text{sum}}}L\alpha^m\right)n^{-\gamma/2} + \sqrt{\lambda_{\text{sum}}}L\alpha^m\|\theta - \hat\theta_n\|\right]\|\theta - \hat\theta_n\| \\
&\quad + \frac{\eta^2}{2}\left[\lambda_{\max}\|\theta - \hat\theta_n\| + \left(1 + \sqrt{\lambda_{\text{sum}}}L\alpha^m\right)n^{-\gamma/2} + \sqrt{\lambda_{\text{sum}}}L\alpha^m\|\theta - \hat\theta_n\|\right]^2 \\
&\quad + \frac{\eta^2}{2} \times \frac{dC_{2,m}}{n}
\end{aligned}$$

(3.6)
$$= -\eta\left(a\|\theta - \hat\theta_n\|^2 - b_n\|\theta - \hat\theta_n\| - c_n\right)$$

whose right-hand side is quadratic in $\|\theta - \hat\theta_n\|$. If $a > 0$ then large $\|\theta - \hat\theta_n\|$ guarantees a negative drift. Specifically,

$$\|\theta - \hat\theta_n\| \geq r_n := \frac{b_n + \sqrt{b_n^2 + 4ac_n}}{2a} \implies \mathbb{E}_\theta^{\mathbf{x}} V(\theta^+) - V(\theta) \leq 0.$$

For any $\beta > 1$, let

$$B := \{\theta \in \Theta : \|\theta - \hat\theta_n\| \leq \beta r_n\}.$$

If $\theta \notin B$ i.e. $\|\theta - \hat\theta_n\| \geq \beta r_n$ then

$$\begin{aligned}
\mathbb{E}_\theta^{\mathbf{x}} V(\theta^+) - V(\theta) &\leq -\eta\left(a\beta^2 r_n^2 - b_n\beta r_n - c_n\right) \\
&= -\eta[a(\beta^2 - 1)r_n^2 - b_n(\beta - 1)r_n] \\
&\leq -\eta(\beta^2 - 1)(ar_n^2 - br_n) \\
&= -\eta(\beta^2 - 1)c_n \\
&= -\delta_1
\end{aligned}$$

On the other hand, if $\theta \in B$,

$$\mathbb{E}_\theta^{\mathbf{x}} V(\theta^+) - V(\theta) \leq \max_{\theta \in \Theta} -\eta \left( a\|\theta - \hat{\theta}_n\|^2 - b_n\|\theta - \hat{\theta}_n\| - c_n \right)$$
$$= \eta(c_n + b_n^2/4a)$$
$$= \delta_2,$$

completing the proof. □

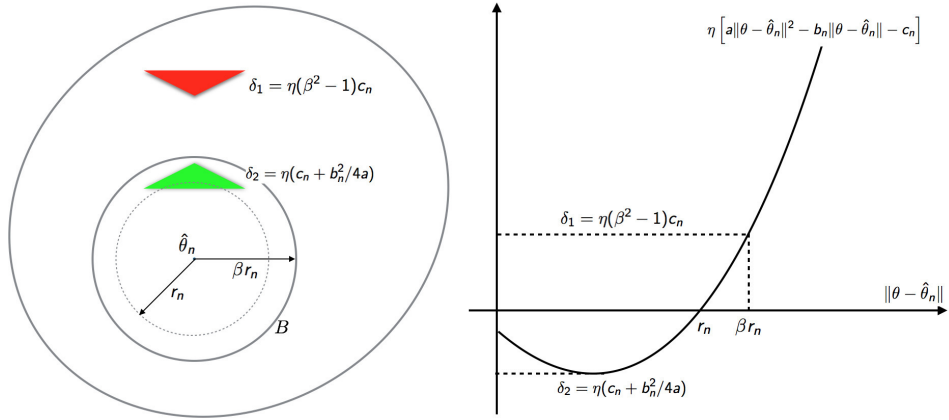

Fig 1: Equation (3.6) asserts that the drift of $V(\theta) = \|\theta - \hat{\theta}_n\|^2/2$ is upper bounded by a quadratic function of $\|\theta - \hat{\theta}_n\|$. It implies the drift criterion that $V$ decreases at least $\delta_1$ after a move from $B^c$, and increases at most $\delta_2$ after a move from $B$.

Figure 1 illustrates the intuition of Lemma 3.3 that the drift of $V(\theta) = \|\theta - \hat{\theta}_n\|^2/2$ is upper bounded by a quadratic function of $\|\theta - \hat{\theta}_n\|$. This bound later implies the drift criterion that $V$ decreases at least $\delta_1$ after a move from $B^c$, and increases at most $\delta_2$ after a move from $B$.

3.4. *Positive Recurrence around the MLE under the Chain.* Lemma 3.4 follows to show that $B$ is positive recurrent under the chain $\{\theta_t\}_{t \geq 0}$ in the sense that the proportion of time that the chain stays inside of $B$ in the long term is at least $\delta_1/(\delta_1 + \delta_2)$. The proof of Lemma 3.4 uses the Azuma-Hoeffding inequality [42] for super-martingales with bounded differences.

LEMMA 3.4. *Following Lemma 3.3, $B = \{\theta \in \Theta : \|\theta - \hat{\theta}_n\| \leq \beta r_n\}$ is positive recurrent under the chain $\{\theta_t\}_{t \geq 0}$ in the sense that*

$$\liminf_{t \to \infty} \frac{1}{t} \sum_{s=0}^{t-1} \mathbb{I}(\theta_s \in B) \geq \frac{\delta_1}{\delta_1 + \delta_2} = \frac{\beta^2 - 1}{\beta^2 + b_n^2/4ac_n}$$

$$\limsup_{t \to \infty} \frac{1}{t} \sum_{s=0}^{t-1} \mathbb{I}(\theta_s \notin B) \leq \frac{\delta_2}{\delta_1 + \delta_2} = \frac{1 + b_n^2/4ac_n}{\beta^2 + b_n^2/4ac_n}$$

PROOF. First construct two super-martingales with bounded differences. Let

$$Y_t = V(\theta_{t+1}) - V(\theta_t) + \delta_1, \quad Z_t = V(\theta_{t+1}) - V(\theta_t) - \delta_2.$$

By the Foster-Lyapunov drift criterion established in Lemma 3.3 and the Markov property of $\{\theta_t\}_{t \geq 0}$,

$$\sum_{s=0}^{t-1} Y_s \mathbb{I}(\theta_s \notin B), \quad \sum_{s=0}^{t-1} Z_s \mathbb{I}(\theta_s \in B)$$

are super-martingales under $\mathbb{P}^{\mathbf{x}}$. They have bounded differences $|Y_t| < D, |Z_t| < D$ for some $D$ since $V(\theta)$ is bounded on compact $\Theta$. By the Azuma-Hoeffding inequality (Lemma 3.5 in the Appendix), for any $\epsilon > 0$

$$\mathbb{P}^{\mathbf{x}}\left(\frac{1}{t} \sum_{s=0}^{t-1} Y_s \mathbb{I}(\theta_s \notin B) \geq \epsilon\right) = \mathbb{P}^{\mathbf{x}}\left(\sum_{s=0}^{t-1} Y_s \mathbb{I}(\theta_s \notin B) \geq t\epsilon\right) \leq \exp\left(-\frac{t\epsilon^2}{2D^2}\right).$$

By Boral-Cantelli Lemma, $\sum_{t=0}^{\infty} \exp\left(-\frac{t\epsilon^2}{2D^2}\right) < \infty$ implies

$$\mathbb{P}^{\mathbf{x}}\left(\frac{1}{t} \sum_{s=0}^{t-1} Y_s \mathbb{I}(\theta_s \notin B) \geq \epsilon \text{ finitely often}\right) = 0.$$

That is,

$$\limsup_{t \to \infty} \frac{1}{t} \sum_{s=0}^{t-1} Y_s \mathbb{I}(\theta_s \notin B) \leq \epsilon \quad \mathbb{P}^{\mathbf{x}}\text{-a.s..}$$

$\epsilon > 0$ can be arbitrarily small. It follows that

$$\limsup_{t \to \infty} \frac{1}{t} \sum_{s=0}^{t-1} Y_s \mathbb{I}(\theta_s \notin B) \leq 0 \quad \mathbb{P}^{\mathbf{x}}\text{-a.s..}$$

Similarly,
$$\limsup_{t\to\infty} \frac{1}{t}\sum_{s=0}^{t-1} Z_s \mathbb{I}\left(\theta_s \in B\right) \leq 0 \quad \mathbb{P}^{\mathbf{x}}\text{-a.s..}$$

Combining them with the fact that
$$\frac{1}{t}\sum_{s=0}^{t-1}(\delta_1 + \delta_2)\mathbb{I}\left(\theta_s \in B\right) = \frac{1}{t}\sum_{s=0}^{t-1}(Y_s - Z_s)\mathbb{I}\left(\theta_s \in B\right)$$
$$= \frac{1}{t}\sum_{s=0}^{t-1} Y_s - \frac{1}{t}\sum_{s=0}^{t-1} Y_s \mathbb{I}\left(\theta_s \notin B\right) - \frac{1}{t}\sum_{s=0}^{t-1} Z_s \mathbb{I}\left(\theta_s \in B\right)$$

yields
$$\liminf_{t\to\infty} \frac{1}{t}\sum_{s=0}^{t-1} \mathbb{I}\left(\theta_s \in B\right) \geq \liminf_{t\to\infty} \frac{1}{(\delta_1 + \delta_2)t} \sum_{s=0}^{t-1} Y_s$$
$$= \liminf_{t\to\infty} \frac{V(\theta_t) - V(\theta_0) + \delta_1 t}{(\delta_1 + \delta_2)t}$$
$$= \frac{\delta_1}{\delta_1 + \delta_2} = \frac{\beta^2 - 1}{\beta^2 + b_n^2/4ac_n} \quad \mathbb{P}^{\mathbf{x}}\text{-a.s..}$$

It is equivalent to say
$$\limsup_{t\to\infty} \frac{1}{t}\sum_{s=0}^{t-1} \mathbb{I}\left(\theta_s \notin B\right) \leq \frac{\delta_2}{\delta_1 + \delta_2} = \frac{1 + b_n^2/4ac_n}{\beta^2 + b_n^2/4ac_n} \quad \mathbb{P}^{\mathbf{x}}\text{-a.s.,}$$

completing the proof. □

This lemma shows that the proportion of time that the chain stays inside of $B$ in the long term is at least
$$\liminf_{t\to\infty} \frac{1}{t}\sum_{s=0}^{t-1} \mathbb{I}\left(\theta_s \in B\right) \geq \frac{\delta_1}{\delta_1 + \delta_2} = \frac{\beta^2 - 1}{\beta^2 + b_n^2/4ac_n} = \frac{\beta^2 - 1}{\beta^2 + \mathcal{O}(1)}.$$

Note that the radius of the closed ball is $\beta r_n$. Letting $\beta \asymp n^{\gamma'/2}$ for any $\gamma' \in (0, \gamma)$, we have
$$\liminf_{t\to\infty} \frac{1}{t}\sum_{s=0}^{t-1} \mathbb{I}\left(\|\theta_s - \hat{\theta}_n\| \leq n^{-(\gamma-\gamma')/2}\right) \geq 1 - \mathcal{O}(n^{-\gamma'})$$

for sufficiently large $n$. As $n \to \infty$, the chain will gradually concentrate at the MLE. Choosing an appropriate $\gamma'$, Lemma 3.5 shows that every limit point of the time average $\frac{1}{t}\sum_{s=0}^{t-1} \theta_s$ is $\mathcal{O}(1/\sqrt[3]{n})$-close to the MLE $\hat{\theta}_n$.

LEMMA 3.5. *Following Lemma 3.4,*

$$\limsup_{t\to\infty} \left\| \frac{1}{t} \sum_{s=0}^{t-1} \theta_s - \hat{\theta}_n \right\| = \mathcal{O}(n^{-\gamma/3}) \quad \mathbb{P}^{\mathbf{x}}\text{-}a.s..$$

PROOF. Write

$$\left\| \frac{1}{t} \sum_{s=0}^{t-1} \theta_s - \hat{\theta}_n \right\| \leq \frac{1}{t} \sum_{s=0}^{t-1} \|\theta_s - \hat{\theta}_n\|$$

$$= \frac{1}{t} \sum_{s=0}^{t-1} \|\theta_s - \hat{\theta}_n\| \mathbb{I}\,(\theta_s \in B) + \frac{1}{t} \sum_{s=0}^{t-1} \|\theta_s - \hat{\theta}_n\| \mathbb{I}\,(\theta_s \notin B)$$

$$\leq \beta r_n + \max_{\theta,\theta'\in\Theta} \|\theta - \theta'\| \times \frac{1}{t} \sum_{s=0}^{t-1} \mathbb{I}\,(\theta_s \notin B)$$

where the first step is due to the convexity of $l_2$-norm. Putting it together with the result

$$\limsup_{t\to\infty} \frac{1}{t} \sum_{s=0}^{t-1} \mathbb{I}\,(\theta_s \notin B) \leq \frac{\delta_2}{\delta_1 + \delta_2} = \frac{1 + b_n^2/4ac_n}{\beta^2 + b_n^2/4ac_n}$$

in Lemma 3.4 yields

$$\limsup_{t\to\infty} \left\| \frac{1}{t} \sum_{s=0}^{t-1} \theta_s - \hat{\theta}_n \right\| \leq \beta r_n + \underbrace{\max_{\theta,\theta'\in\Theta} \|\theta - \theta'\|}_{\text{not depend on } n} \times \frac{1 + b_n^2/4ac_n}{\beta^2 + b_n^2/4ac_n}$$

Recall that $r_n \asymp n^{-\gamma/2}, b_n^2/4ac_n \asymp 1$. If $\beta \asymp n^{\gamma'/2}$ increases with $n$, then

$$\limsup_{t\to\infty} \left\| \frac{1}{t} \sum_{s=0}^{t-1} \theta_s - \hat{\theta}_n \right\| = \mathcal{O}\left(\max\{n^{-(\gamma-\gamma')/2}, n^{-\gamma'}\}\right).$$

The bound is minimized when $\gamma' = \gamma/3$ such that

$$(\gamma - \gamma')/2 = \gamma'.$$

That is,

$$\limsup_{t\to\infty} \left\| \frac{1}{t} \sum_{s=0}^{t-1} \theta_s - \hat{\theta}_n \right\| = \mathcal{O}(n^{-\gamma/3}).$$

The coefficient constant depends on $m$, since $\max_{\theta,\theta'\in\Theta} \|\theta - \theta'\|$ depends on the size of parameter space $\Theta$ only, and the ratio of $b_n^2/4ac_n$ in the limit of $n$ is determined by $m$. □

3.5. *Proof of the Main Theorem.* Now we can complete the proof of the main result in Theorem 2.1.

PROOF OF THEOREM 2.1. In the light of Lemma 3.1, it suffices to show

$$\lim_{n \to \infty} \mathbb{P}^{\mathbf{x}} \left( \limsup_{t \to \infty} \left\| \frac{1}{t} \sum_{s=0}^{t-1} \theta_s - \theta_\star \right\| > A_m n^{-\gamma/3} \right) = 0$$

for any data sample $\mathbf{x}$ satisfying (3.1), (3.2) and (3.3). To this end, Lemma 3.5 asserts that

$$\limsup_{t \to \infty} \left\| \frac{1}{t} \sum_{s=0}^{t-1} \theta_s - \hat{\theta}_n \right\| = \mathcal{O}(n^{-\gamma/3}) \quad \mathbb{P}^{\mathbf{x}}\text{-a.s.}$$

and constraint (3.2) ensures $\|\hat{\theta}_n - \theta_\star\| < n^{-\gamma/2}$. Combining them yields

$$\limsup_{t \to \infty} \left\| \frac{1}{t} \sum_{s=0}^{t-1} \theta_s - \theta_\star \right\| \leq \limsup_{t \to \infty} \left\| \frac{1}{t} \sum_{s=0}^{t-1} \theta_s - \hat{\theta}_n \right\| + \|\hat{\theta}_n - \theta_\star\|$$
$$= \mathcal{O}(n^{-\gamma/3}) \quad \mathbb{P}^{\mathbf{x}}\text{-a.s.}$$

as desired. And the coefficient constant does not depend on $\mathbf{x}$. □

**4. Examples.** We provide three examples: a bivariate Gaussian model, an $2 \times 2$ Fully-visible Boltzmann Machine (FVBM) and an exponential family random graph model to illustrate our theories. For each example, we first verify the assumptions (A1)-(A6) one by one, and then show two phases, namely "quick move" and "random walk", of the CD learning process by plotting the sequence $\{\theta_t\}_{t \geq 0}$.

4.1. *Bivariate Gaussian.* We take a bivariate Gaussian model with unknown mean $\theta \in \mathbb{R}^2$ but known covariance matrix

$$\Sigma = \begin{bmatrix} \sigma_1^2 & \rho\sigma_1\sigma_2 \\ \rho\sigma_1\sigma_2 & \sigma_2^2 \end{bmatrix} = \begin{bmatrix} 1.0 & 0.5 \\ 0.5 & 1.0 \end{bmatrix}.$$

as our first example. We abuse the notations $X^{(1)}$ and $X^{(2)}$ to denote the components of the Gaussian variable $X \sim \mathcal{N}(\theta, \Sigma)$, and $\theta^{(1)}$ and $\theta^{(2)}$ to denote the components of the parameter $\theta$. This Gaussian model is a canonical exponential family with

$$\phi(X) = \Sigma^{-1} X, \quad \Lambda(\theta) = \theta^T \Sigma^{-1} \theta / 2.$$

We run the CD algorithm with a random-scan Gibbs sampler

$$X^{(1)}|X^{(2)} \sim \mathcal{N}\left(\theta^{(1)} + \rho\frac{\sigma_1}{\sigma_2}\left(X^{(2)} - \theta^{(2)}\right), (1-\rho^2)\sigma_1^2\right),$$

$$X^{(2)}|X^{(1)} \sim \mathcal{N}\left(\theta^{(2)} + \rho\frac{\sigma_2}{\sigma_1}\left(X^{(1)} - \theta^{(1)}\right), (1-\rho^2)\sigma_2^2\right).$$

on three data samples of size $n = 10^2, 10^3, 10^4$ for evaluation purpose. The data samples are generated with the true mean parameter $\theta_\star = (0,0)^T \in \Theta = [-0.5, +0.5] \times [-0.5, +0.5]$. For each data sample, CD starts from $\theta_0 = (0.5, 0.5)^T$, and iterates $T = 2000$ times with learning rate $\eta = 0.01$.

Let us first verify (A1)-(A6). (A1) trivially holds since $\theta_\star = (0,0)^T$ is a interior point of $\Theta = [-0.5, +0.5] \times [-0.5, +0.5]$. And for any $\theta \in \Theta$, $\nabla^2 \Lambda(\theta) = \Sigma^{-1}$.

$$\lambda_{\min}(\theta) = \lambda_{\min} = 0.67, \quad \lambda_{\max}(\theta) = \lambda_{\max} = 2.00, \quad \lambda_{\text{sum}}(\theta) = \lambda_{\text{sum}} = 2.67.$$

For (A2), solving an optimization problem yields

$$L = \max_{\theta \in \Theta} \frac{\chi(p_{\theta_\star}, p_\theta)}{\|\theta - \theta_\star\|} = \max_{\theta \in \Theta} \frac{\sqrt{e^{\theta^T \Sigma^{-1} \theta} - 1}}{\|\theta\|} \approx 1.84.$$

For (A3), using the explicit expression of $\alpha(\theta)$ in [32] we have

$$1 - \alpha(\theta) = 1 - \alpha = \text{smallest eigenvalue of } \Sigma^{-1}/d = 0.33.$$

The $m$-step distribution $k_\theta^m(x, \cdot)$ of the chain starting from any $x \in \mathbb{R}^2$ is essentially Gaussian, and

$$f_\theta(x) = \int_{\mathcal{X}} \phi(y) k_\theta^m(x, y) dy = A(m, \Sigma)\theta + B(m, \Sigma)x$$

is linear in $\theta$ and $x$ with coefficient matrices $A$ and $B$ depending on $m$ and $\Sigma$. Thus (A4), (A5), (A6) hold. When $m \geq 4$ and $\eta = 0.01$, condition (2.5) is satisfied.

By our theories, the CD-4 algorithm will generate a sequence $\{\theta_t\}_{t \geq 0}$, which quickly moves to $\hat{\theta}_n$ and then randomly walks around $\hat{\theta}_n$. The range of the random walk decreases as $n$ increases. Figure 2 illustrates this phenomenon by plotting $\{\theta_t\}_{t=0}^{2000}$. Figure 3 plots $\|\theta_t - \hat{\theta}_n\|$ versus iteration $t$ and clearly shows two phases: "quick move" and "random walk" of the CD learning process. The random walk phase starts at $t \approx 500, 750, 1000$ for $n = 10^2, 10^3, 10^4$, respectively. Larger samples result in smaller random walk neighborhoods.

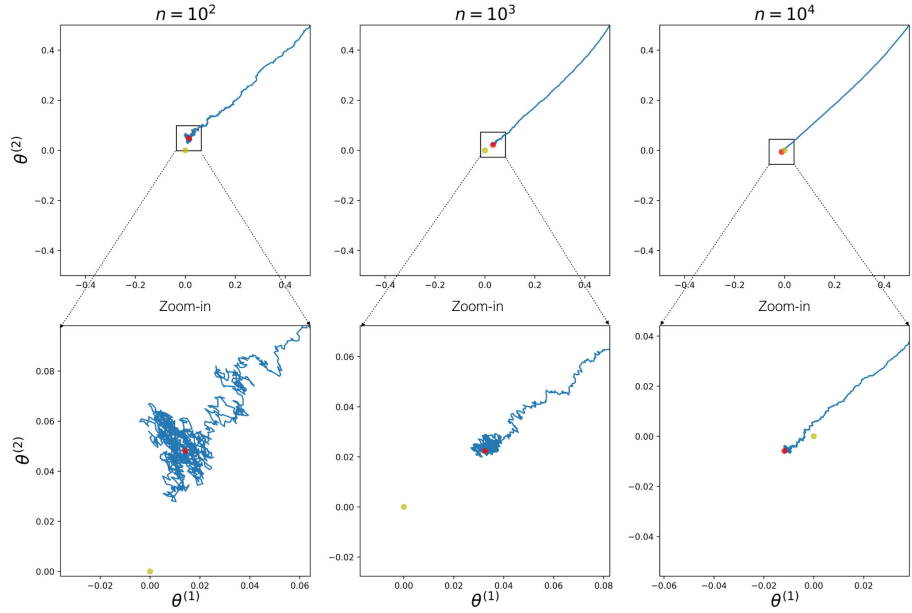

Fig 2: $\{\theta_t\}_{0 \le t \le 2000}$ generated by CD-4 using Gaussian samples of size $n = 10^2$ (left), $10^3$ (middle), $10^4$ (right). Red dots are the MLE $\hat{\theta}_n$, and yellow dots are the true parameter $\theta_\star$.

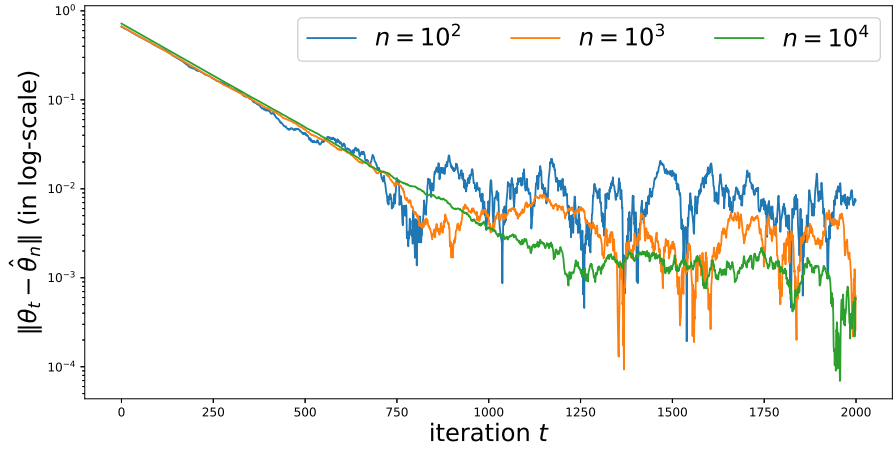

Fig 3: $\|\theta_t - \hat{\theta}_n\|$ versus iteration $t$ for CD-4 using Gaussian samples.

4.2. *Fully-visible Boltzmann Machine.* We analyze CD-1 in $2 \times 2$ FVBM (i.e. $x$ is bivariate) as another example to illustrate our theories. Let $\theta = (b_1, b_2, W_{12}) \in \mathbb{R}^3$, and write $2 \times 2$ FVBM in the canonical form of exponential family

$$p_\theta(x^{(1)}, x^{(2)}) \propto \exp\left(\theta^{(1)} x^{(1)} + \theta^{(2)} x^{(2)} + \theta^{(3)} x^{(1)} x^{(2)}\right).$$

Here we abuse the notations $X^{(1)}$ and $X^{(2)}$ to denote the components of the FVBM variable $X$, and $\theta^{(1)}, \theta^{(2)}, \theta^{(3)}$ to denote the components of the parameter $\theta$. The data samples are generated with the true parameter $\theta_\star = (0, 0, 0.5)^T \in \Theta = [-1, +1] \times [-1, +1] \times [-0.5, +1.5]$. For each data sample, CD-1 starts from $\theta_0 = (1, 1, 1)^T$, and iterates $T = 2000$ times with learning rate $\eta = 0.01$. Each iteration uses a random-scan Gibbs sampler

$$X^{(1)}|X^{(2)} \sim 2 \times \text{Bernoulli}\left(p_\theta(X^{(1)} = +1|X^{(2)})\right) - 1$$
$$X^{(2)}|X^{(1)} \sim 2 \times \text{Bernoulli}\left(p_\theta(X^{(2)} = +1|X^{(1)})\right) - 1$$

We can verify (A1)-(A6) one by one. (A1) holds since $\theta_\star$ is an interior point of $\Theta$. (A2) holds since $\Lambda(\theta) < \infty$ for any $\theta \in \mathbb{R}^3$. For (A3), $\alpha(\theta)$ is the second largest absolute eigenvalue of the transition probability matrix, which is less than 1 and continuous in $\theta \in$ compact $\Theta$, and thus has an upper bound $\alpha < 1$. (A4) and (A6) hold because components of $\phi(X^{(m)})$ are bounded random variables. $k_\theta^m(x, y)$ can be represented as a transition probability matrix whose entries are continuously differentiable functions of $\theta$. Then for any $x \in \mathcal{X} = \{-1, +1\}^p$, $f_\theta(x)$ is continuously differentiable and thus Lipschitz continuous in $\theta \in$ compact $\Theta$. In addition, $\mathcal{X}$ is a finite set, thus (A5) holds.

Figure 4 plots the sequence $\{\theta_t\}_{t=0}^{2000}$ generated by CD-1 on FVBM samples. At the beginning, $\theta_t$ moves quickly towards the MLE and then randomly walks around it. The range of the random walk decreases as $n$ increases.

4.3. *Exponential-family Random Graph Model.* We analyze CD-5 in exponential family random graph model (ERGM) with 10 nodes as the third example. ERGM is widely used in social network analysis [43]. The CD algorithm has performed well in these models [15, 16, 17].

Assume we have a undirected graph $x$ with $x_{ij} = 1$ indicating the existence of an edge between $i$-th node and $j$-th node, and $x_{ij} = 0$ otherwise. The probability mass function is given by $p_\theta(x) \propto \exp[\theta^T \phi(x)]$, where $\phi(x) = [\phi_1(x), ..., \phi_d(x)]$ are the global features of the network. Like [16], our experiment sets $\phi(x)$ to be three network statistics, namely, the number of edges, the number of stars, and the number of triangles. The data samples are generated with the true parameter $\theta_\star = (-2, 0, 0)^T \in \Theta = [-5, +5] \times [-5, +5] \times [-5, +5]$. For each data sample, CD-5 starts from $\theta_0 = (5, 5, 5)^T$, and iterates $T = 500$ times with learning rate $\eta = 0.05$. Each iteration uses a Metropolis-Hastings sampler.

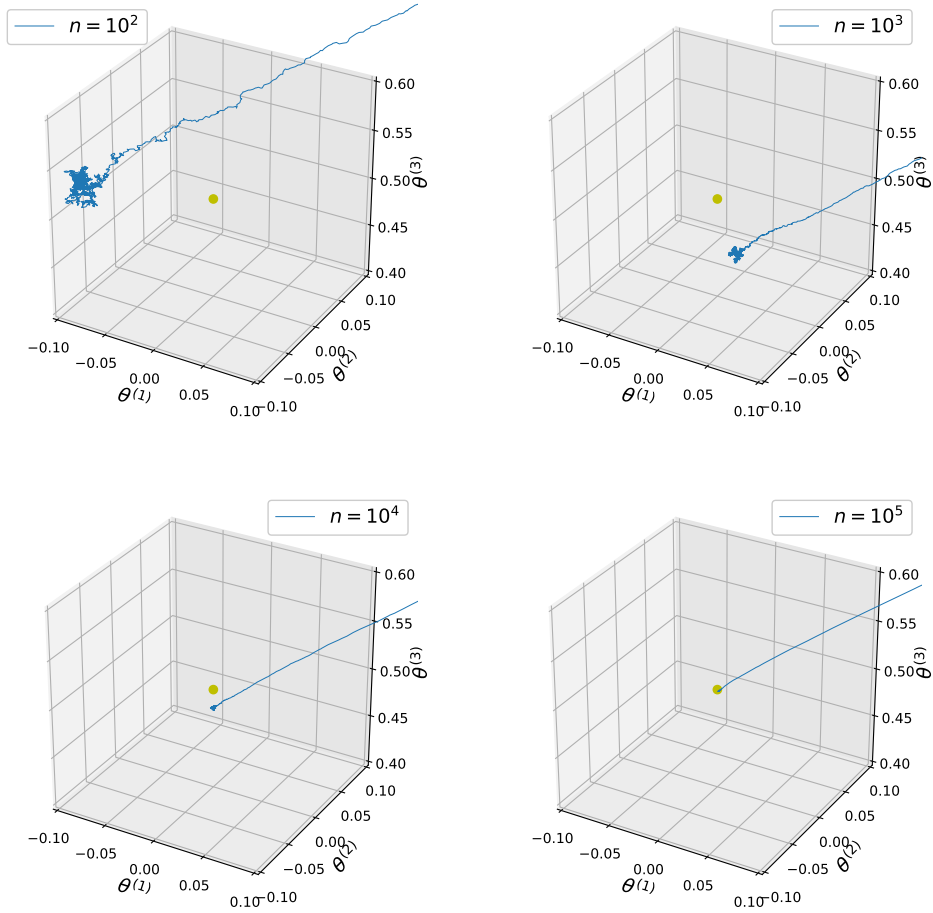

Fig 4: For each FVBM sample of size $n = 10^2$ (upper-left), $10^3$ (upper-right), $10^4$ (lower-left), $10^5$ (lower-right), CD-1 generates $\{\theta_t\}_{0 \leq t \leq 2000}$. The yellow dots are the true parameter $\theta_\star = (0, 0, 0.5)^T$.

We can verify (A1)-(A6) one by one. (A1) holds since $\theta_\star$ is an interior point of $\Theta$. (A2) holds since $\Lambda(\theta) < \infty$ for any $\theta \in \mathbb{R}^3$. For (A3), $\alpha(\theta)$ is the second largest absolute eigenvalue of the transition probability matrix, Each entry of the transition probability matrix is a continuous function in $\theta \in$ compact $\Theta$. It follows that $\alpha(\theta)$ is less than 1 and continuous in $\theta \in$ compact $\Theta$. Thus it has an upper bound $\alpha < 1$. (A4) and (A6) hold because components of $\phi(X^{(m)})$ are bounded. Then for any $x \in \mathcal{X} = \{0,1\}^{10}$, $f_\theta(x)$ is continuously differentiable and thus Lipschitz continuous in $\theta \in$ compact $\Theta$. In addition, the sample space $\mathcal{X}$ is a finite set, thus (A5) holds.

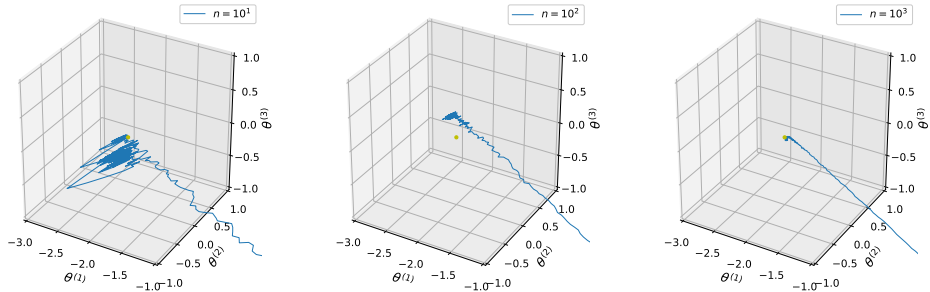

Fig 5: For each ERGM sample of size $n = 10^1$ (left), $10^2$ (middle) and $10^3$ (right), CD-5 generates $\{\theta_t\}_{t=0}^{500}$. The yellow dots are the true parameter $\theta_\star = (-2, 0, 0)^T$.

Figure 5 plots the sequence $\{\theta_t\}_{t=0}^{500}$ generated by CD-5 on ERGM samples. At the beginning, $\theta_t$ moves quickly towards the MLE and then randomly walks around it. The range of the random walk decreases as $n$ increases.

**5. Discussion.** The CD algorithm was proposed to train energy-based models (e.g. Restricted Boltzmann Machine) and later used in the layerwise pre-training step of deep belief network. It has played a key role in the emergence of deep learning. The algorithm approximates the gradient of the log-likelihood function by running short MCMC chains to save computational cost. Although using biased gradient approximations in the iteration of the gradient ascent update, the CD algorithm gives satisfactory parameter estimates in many practical applications. On the other hand, MacKay [18] provides Gaussian examples in which the CD-1 algorithm does not converge to the MLE. Many eminent scholars in machine learning including Yuille [23], Carreira-Perpinan and Hinton [20], Bengio and Delalleau [21] and Sutskever and Tieleman [22] had attempted to theoretically analyze the CD algorithm. But whether and why this algorithm is asymptotically consistent is still an open question.

In order to fill in the gap, this paper is devoted to a theoretical analysis of the CD algorithm in exponential families. Exponential families are special cases of the energy-based models with convex log-partition functions. We narrow the scope of analyses down to these special cases, because (1) the convexity ensures the uniqueness of the MLE so that we could compare the CD estimates to the unique MLE, and (2) the CD algorithm has been used to fit exponential families with unavailable gradient of log-partition function e.g. exponential-family random graph models. For non-convex energy-based models like Restricted Boltzmann Machines, we conjecture that the CD algorithm converges to a random walk around one or more local maximum points of the likelihood function. Intuitively, the gradient ascent algorithm using exact gradients usually does not converge to the MLEs of non-convex models. We do not expect the CD algorithm using approximate gradients does better than the gradient ascent algorithm.

We find that $\{\theta_t\}_{t\geq 0}$ is a homogenous Markov chain conditional on the data sample. And the chain quickly moves towards the MLE $\hat{\theta}_n$ and then randomly walk around it. This phase transition can be explained as follows. When the chain moves from $\theta$ which is far away from the MLE $\hat{\theta}_n$, the exact gradient $g(\theta)$ is relatively large compared to the approximation error resulting from the $m$-step MCMC sampling. The CD update equation keeps pushing $\theta$ to quickly move towards $\hat{\theta}_n$. When $\theta$ enters a small neighbor around $\hat{\theta}_n$, $g(\theta)$ fails to suppress the MCMC approximation error. The "quick move" phase ends and the chain starts a "random walk" in this neighborhood. In addition to the plots of the sequence $\{\theta_t\}_{t\geq 0}$ in Section 4, we plot the gradient fields of the bivariate Gaussian and the FVBM examples and put them in the Appendix.

Our theories can explain MacKay [18]'s Gaussian examples in which the CD-1 algorithm does not converge to the MLE. Mackay found that the CD-1 algorithm with different kernels has one or multiple fixed points but these fixed points may not be the MLEs. However, the author reported that "in the special case of an infinite amount of data that come from an axis-aligned Gaussian, the noisy swirl operator's one-step algorithm (CD-1) does converge to the maximum-likelihood parameters." According to our theories, the fixed points or the limit points of the sequence $\{\theta_t\}_{t\geq 0}$ are not exactly the MLEs. But they are $\mathcal{O}(1/\sqrt[3]{n})$-close to MLEs with high probability and the gap shrinks to 0 as the amount of data goes to infinity. In this way, our theories give an explanation for Mackay's Gaussian examples. We think the phenomenon that CD does not converge to the MLE should not be considered as a failure of CD. We note that the $\mathcal{O}_p(1/\sqrt[3]{n})$ rate is probably not the best rate. We believe that a rate closer to or equals to $\mathcal{O}_p(1/\sqrt{n})$ should be obtainable with a more refined analysis.

We would like to highlight three other novelties of the theoretical analyses in this paper. First, we let the iteration number $t \to \infty$ and then let $n \to \infty$ when analyzing the CD algorithm, while many previous work had not clearly distinguish the two limits and led to unreasonable convergence conditions. For example, two convergence conditions given in the remarks of Result 4 in [23] (translated to the form using our notations) are

$$\frac{1}{n}\sum_{i=1}^{n} \phi(X_i) = \nabla\Lambda(\theta_\star)$$

$$\frac{1}{n}\sum_{i=1}^{n}\int_{\mathcal{X}} \phi(y) k_{\theta_\star}^m(X_i, y) dy = \nabla\Lambda(\theta_\star).$$

These conditions are satisfied with probability zero in most models of continuous distributions. It is because their left-hand-sides are functions of $\{X_i\}_{i=1}^n$ and thus random, while their right-hand-sides are non-random. Second, conditioning on a finite data sample, we can see the sequence $\{\theta_t\}_{t\geq 0}$ as a Markov chain. Hence, many nice results in the Markov chain, martingale and stability analysis theories like Foster-Lyapunov drift criterion can be used. We believe Lyapunov drift conditions might be used to analyze other stochastic gradient descent schemes with approximate gradients. Third, the sub-exponentiality assumption (A4) allows to analyze models with unbounded sufficient statistics. It is an improvement over existing theoretical analyses, which are restricted to bounded cases.

Our theoretical results also provide some guidance for the practitioners of the CD algorithm. First, since the CD algorithm converges to a random walk around the MLE, a single estimate $\theta_t$ is not as reliable as the average $\frac{1}{t}\sum_{s=0}^{t-1}\theta_s$. Thus an averaging scheme should be taken. Second, the CD learning process typically has two phases: "quick move" and "random walk". This phase division suggests practitioners to stop the iterations of the algorithm when $\theta_t$ starts the random walk. Third, the success of the CD algorithm highly relies on the speed of MCMC kernels in use. One may design and test a few candidate MCMC kernels, and use the fastest kernel in the CD algorithm. Last but not the least, one could do mini-batch sampling at each iteration of the CD algorithm, as a smaller sample size like $n/10$ would change neither the bias of the CD gradient approximation nor the order of its variance. Thus the asymptotical consistency of the CD algorithm still holds if the mini-batch sampling scheme is in use.

## SUPPLEMENTARY MATERIAL

**Appendix: Other Simulation Results and Lemmas**
(doi: COMPLETED BY THE TYPESETTER; .pdf). This supplementary material contains other simulation results and five lemmas.

**Acknowledgements.** This work is supported by NSF of US under grant DMS-1407557. The authors would like to thank Dr. Weijie Su, Dr. Rachel Wang, Dr. Lester Mackey and Dr. Percy Liang for valuable advice.

BAI JIANG, TUNG-YU WU, YIFAN JIN
WONG LAB, JAMES H. CLARK CENTER, E100
318 CAMPUS DRIVE
STANFORD UNIVERSITY, CA 94305
E-MAIL: baijiang@stanford.edu
    tungyuwu@stanford.edu
    yifanj@stanford.edu

WING H. WONG
DEPARTMENT OF STATISTICS
390 SERRA MALL
STANFORD, CA 94305
E-MAIL: whwong@stanford.edu
URL: http://web.stanford.edu/group/wonglab/